\documentclass[final,3p]{elsarticle1}

\usepackage{lineno}
\usepackage[colorlinks=true,breaklinks=true,pdftex]{hyperref}
\modulolinenumbers[1]
\usepackage{algorithm}
\usepackage{algorithmic}
\usepackage{amsmath,amsthm,amssymb,amsfonts}
\usepackage{cleveref}  

\usepackage{subfigure}
\usepackage{amssymb}
\usepackage{multirow}
\crefname{section}{Sec.}{Secs.}
\Crefname{section}{Section}{Sections}
\crefname{table}{Tab.}{Tabs.}
\Crefname{table}{Table}{Tables}
\crefname{figure}{Fig.}{Figs.}
\Crefname{figure}{Figure}{Figures}
\crefname{equation}{Eq.}{Eqs.}
\Crefname{equation}{Equation}{Equations}
\crefname{theorem}{Thm.}{Thms.}
\Crefname{theorem}{Theorem}{Theorems}
\crefname{lemma}{Lem.}{Lems.}
\Crefname{lemma}{Lemma}{Lemmas}
\crefname{corollary}{Cor.}{Cors.}
\Crefname{corollary}{Corollary}{Corollaries}
\crefname{algorithm}{Alg.}{Algs.}
\Crefname{algorithm}{Algorithm}{Algorithms}

\newcommand{\tocite}[1]{\textcolor{red}{[TO CITE]}}

\usepackage{color}
\definecolor{Red}{cmyk}{0,1,1,0}
\definecolor{Green}{cmyk}{1,0,1,0}
\definecolor{Cyan}{cmyk}{1,0,0,0}
\definecolor{Purple}{cmyk}{0.45,0.86,0,0}
\definecolor{Rosolic}{cmyk}{0.00,1.00,0.50,0}
\definecolor{Blue}{cmyk}{1.00,1.00,0.00,0}
\definecolor{Orange}{cmyk}{0,0.52,0.80,0}
\definecolor{Black}{cmyk}{1,0,0,1}

\biboptions{authoryear}

\begin{document}

\begin{frontmatter}

\title{Automatic Tooth Arrangement with Joint Features of  Point and Mesh Representations via Diffusion Probabilistic Models}


\author[first]{Changsong Lei}
\ead{leics23@tsinghua.edu.cn}
\author[first]{Mengfei Xia}
\ead{xmf20@tsinghua.edu.cn}
\author[second]{Shaofeng Wang}
\ead{2939108747@ccmu.edu.cn}
\author[first]{Yaqian Liang\corref{cor}}
\ead{yaqianliang@tsinghua.edu.cn}
\author[third]{Ran Yi}
\ead{ranyi@sjtu.edu.cn}
\author[four]{Yu-Hui Wen}
\ead{yhwen1@bjtu.edu.cn}
\author[first]{Yong-Jin Liu\corref{cor}}
\ead{liuyongjin@tsinghua.edu.cn}
\cortext[cor]{Corresponding author}
\address[first]{Department of Computer Science and Technology, Tsinghua University, China}
\address[second]{Department of Orthodontics, Beijing Stomatological Hospital, Capital Medical University, Beijing, China}
\address[third]{School of Electronic Information and Electrical Engineering, Shanghai Jiao Tong University, China}
\address[four]{School of Computer and Information Technology, Beijing Jiaotong University, China}

\begin{abstract}
Tooth arrangement is a crucial step in orthodontics treatment, in which aligning teeth could improve overall well-being, enhance facial aesthetics, and boost self-confidence.  
To improve the efficiency of tooth arrangement and minimize errors associated with unreasonable designs by inexperienced practitioners, some deep learning-based tooth arrangement methods have been proposed.
Currently, most existing approaches employ MLPs to model the nonlinear relationship between tooth features and transformation matrices to achieve tooth arrangement automatically. However, the limited datasets (which to our knowledge, have not been made public) collected from clinical practice constrain the applicability of existing methods, making them inadequate for addressing {\bf diverse} malocclusion issues. To address this challenge, we propose a general tooth arrangement neural network based on the diffusion probabilistic model. Conditioned on the features extracted from the dental model, the diffusion probabilistic model can learn the distribution of teeth transformation matrices from malocclusion to normal occlusion by gradually denoising from a random variable, thus more adeptly managing real orthodontic data. To take full advantage of effective features, we exploit both mesh and point cloud representations by designing different encoding networks to extract the tooth (local) and jaw (global) features, respectively. In addition to traditional metrics ADD, PA-ADD, CSA, and $\mathrm{ME}_{rot}$, we propose {\bf a new evaluation metric} based on dental arch curves to judge whether the generated teeth meet the individual normal occlusion. Experimental results demonstrate that our proposed method achieves state-of-the-art tooth alignment results and satisfactory occlusal relationships between dental arches. We will publish the code and dataset.

\end{abstract}


\end{frontmatter}


 
\section{Introduction}


Malocclusion refers to irregular tooth arrangement, the abnormal relationship between the upper and lower dental arches, or abnormal size/shape/position of the jaws~\citep{wsf5}. 
Currently, a significant number of people worldwide are 
plagued by malocclusion, with the probability of malocclusion ranges
from 39\% to 93\% depending on region and ethnicity~\citep{wsf2,wsf3,wsf4}. Orthodontic treatment is the primary approach to address malocclusion. By applying specific forces in different directions with various types of orthodontic appliances, teeth can be moved towards predetermined positions to achieve an aesthetically pleasing,  balanced, and stable dental arch and occlusal relationship~\citep{wsf6}. 
\begin{figure*}[t]
	\addtocounter{figure}{-1} 
	\centering
	\subfigure{
		\begin{minipage}[t]{0.5\linewidth}
	\includegraphics[width=1\linewidth]{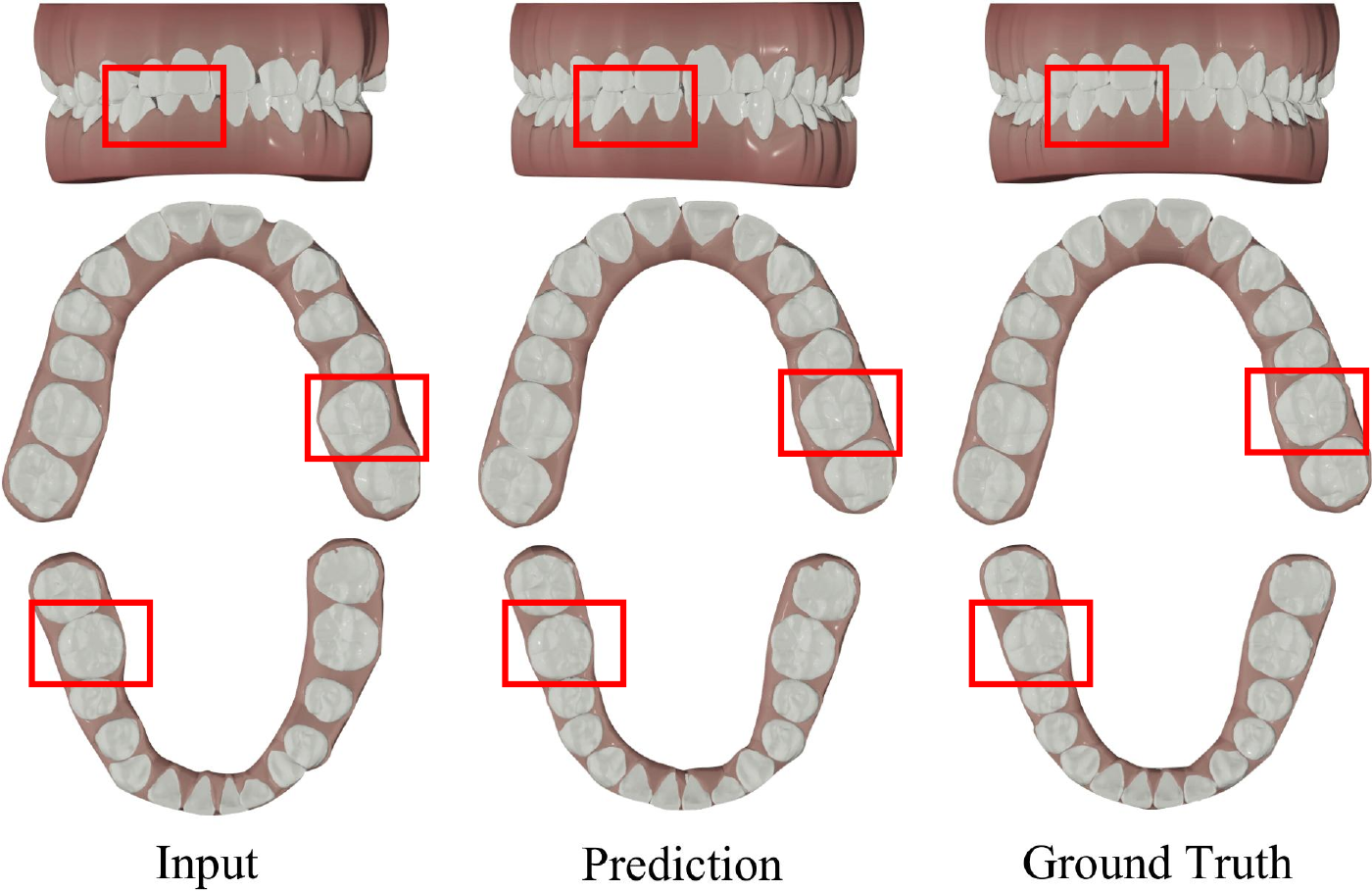}
		\caption{Comparison between the input dental model,  the predicted result of our method, and the ground truth. In this paper,
    we focus on not only the teeth alignment but also the occlusion between the upper and lower dental arches.}
		\label{fig:first}
		\end{minipage}%
	} 
 \hspace{10mm}
	\subfigure{
		\begin{minipage}[t]{0.4\linewidth}
	\includegraphics[width=0.9\linewidth]{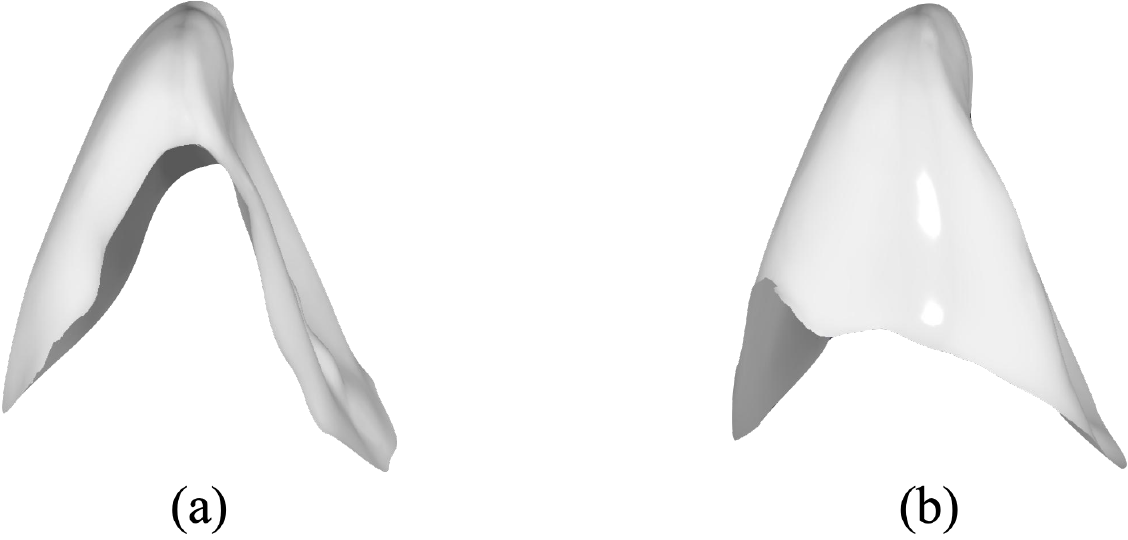}
			\caption{A single tooth in our dataset, where (a) and (b) are the same tooth before and after orthodontic treatment observed from the same angle.}
			\label{fig:single}
		\end{minipage}%
	}%
 
	\centering
\end{figure*}
\addtocounter{figure}{1}

Traditional orthodontic treatment methods require dentists and aligner design technicians to manually arrange teeth on digital dental models. Although current computer-aided design systems have simplified this task to some extent, the process still involves extensive and time-consuming human-computer interaction.
With the growing integration of digital technology in orthodontic practices, digital tooth alignment simulations are becoming more prevalent in clinical settings~\citep{wsf8}. These tools assist dentists in creating treatment plans by simulating teeth movements and predicting their final optimal positions.
In recent years, the successful applications of artificial intelligence technology have markedly propelled the advancement of the medical field. 
Along with it, a variety of data-driven methods have been proposed in dentistry and orthodontics, such as tooth segmentation on oral CBCT data~\citep{li2022semantic, cui2022fully} and teeth alignment prediction
~\citep{wei2020tanet,lingchen2020iorthopredictor,wang2022tooth}. 

Existing tooth arrangement network models rely on the post-treatment tooth segmentations as the ground truth to calculate the loss functions, making tooth segmentation an essential step. 
However, due to the possibility of changes in crown shape during clinical treatment, we found that tooth segmentation results, whether from state-of-the-art intelligent methods or manually annotated segmentations, are not always completely consistent before and after orthodontic treatment. Fig.~\ref{fig:single} demonstrates the segmentation results of a single tooth. Although Fig.~\ref{fig:single}(a) and Fig.~\ref{fig:single}(b) are the same tooth before and after orthodontic treatment, there are significant differences between their shapes.
This kind of inconsistency will affect the calculation of the loss functions, e.g., (1) the chamfer distance between the predicted teeth and ground truth teeth, and (2) the difference between the predicted teeth transformation matrix and the ground truth teeth transformation matrix. The aforementioned issue leads to the existing methods yielding inaccurate predictions of transformation matrices.

To tackle this challenge,  we design our method with two steps: 
(1) We construct a simulated pre-orthodontic teeth dataset by randomly shifting the teeth after orthodontic treatment, thereby ensuring that each tooth segmentation is exactly the same {\it before} and {\it after} orthodontic treatment. 
At the same time, we note that due to large disparities between the distribution of the constructed pre-orthodontic dataset and the distribution of the real dataset,
directly training existing methods on the constructed dataset still results in poor generalization for real orthodontic cases.
(2) To better cope with real orthodontic data, we further propose to use the diffusion model to learn the real distribution of teeth's transformation matrices when moving from an unreasonable (i.e., malocclusion) position to a reasonable (i.e., individual normal occlusion) position. Diffusion probabilistic models (DPMs)~\citep{ho2020denoising} have recently received growing attention due to their powerful representation learning ability and effectiveness in modeling complex data distributions.
We utilize the features extracted from the dental model as the condition guiding the diffusion model, where the 3D mesh models are utilized to extract the local geometric details of a single tooth, and the point clouds are utilized to extract the global shape of the whole dental model. Compared with existing methods, our two-step framework demonstrates superior generalization to real patient data.




To sum up, we make the following contributions:
\begin{itemize}
    \item We propose TADPM, an automatic Tooth Arrangement neural network via Diffusion Probabilistic Model. By generating the teeth transformation matrices and applying them on the teeth models, our method predicts satisfactory orthodontic effects.
    \item To extract effective dental model features, we design different encoder networks at local and global levels. 

     \item Integrating professional orthodontic knowledge, we propose a new  metric based on dental arch curves to precisely evaluate the occlusal relationship between dental arches. We apply it to validate the alignment performance of our tooth arrangement network.


\end{itemize}
Experiments based on a real dataset collected from patients
show that compared to existing methods, our method can significantly improve the teeth alignment results.



\section{Related Works}
\subsection{Learning-based Teeth Processing Methods}

The two most critical steps in orthodontic treatment are segmenting the 3D dental model and developing the arrangement plan. Manually performing these two tasks is time-consuming, tedious, and highly dependent on orthodontists’ experiences due to the abnormality and large-scale variance of patients’ teeth. In recent years, the successful applications of artificial intelligence technology have provided a strong impetus for advancements in various medical fields~\citep{xu20183d}.
For example, early works propose to use neural networks to conduct tooth segmentation on CBCT images~\citep{wirtz2018automatic,cui2019toothnet,shaheen2021novel,cui2022fully,xie2023wits,li2022semantic} or X-ray images~\citep{zhao2020tsasnet}. 
To address problems caused by the irregular data format of 3D models, some works introduce Graph Convolution Network (GCN) to learn discriminative geometric features for 3D dental model segmentation~\citep{sun2020automatic,zhang2021tsgcnet,zheng2023teethgnn}. 
To alleviate the issue of insufficient annotations on 3D dental model data, 
\cite{qiu2022darch} proposes a dental arch prior-assisted 3D tooth segmentation method when only weak annotation is provided, and 
\cite{wu2022two} proposes a two-stage framework based on mesh deep learning (called TS-MDL) for joint tooth labeling and landmark identification on raw intraoral scans.

Tooth arrangement methods are usually developed based on segmentation results.
TANet~\citep{wei2020tanet} is the first learning-based tooth arrangement predicting approach, which employs PointNet to extract the features of the crown point cloud and utilizes a graph neural network (GNN) to implement feature propagation between teeth through topological relations. Iorthopredictor~\citep{lingchen2020iorthopredictor} generates the facial image with aligned teeth, mimicking a real orthodontic treatment effect. 
\cite{li2020malocclusion} proposes to conduct tooth arrangement by leveraging the spatial interrelationship between different teeth. 
\cite{wang2022tooth} proposes a tooth arrangement network based on tooth landmark constraints and a hierarchical graph structure. 
All the aforementioned methods employ point clouds as input data. Nevertheless, the discrete points sampled from dental models lose geometric details and topological connections among teeth, thus potentially impacting subsequent transformation prediction processes.
In contrast, we propose to study automatic tooth arrangement methods with 3D dental mesh models, which is unexplored. Experiments show that extracting information from mesh models with more geometric details yields better results. 

\subsection{Diffusion Probabilistic Models}

Diffusion probabilistic models (DPMs) are a class of generative models that transform a Gaussian distribution into the distribution of the given data via an iterative denoising process~\citep{sohl2015deep,song2019generative,song2020score,ho2020denoising}.
Suppose that $\mathbf x_0\in\mathbb R^D$ is a $D$-dimensional random variable with an unknown distribution $q_0(\mathbf x_0)$.
DPMs define a forward diffuse process by
gradually corrupting the information of $\mathbf x_0$ with Gaussian noises, such that for any 
$t\in[0,T]$, the transition distribution is~\citep{song2020score,sohl2015deep}:
\begin{align}\label{eq:forward}
q_{0t}(\mathbf x_t|\mathbf x_0)=\mathcal N(\mathbf x_t;\alpha_t\mathbf x_0,\sigma_t^2\mathbf I),
\end{align}
where $\alpha_t$ and $\sigma_t>0$ are differentiable functions of $t$ with bounded derivatives. 
%
%


To date, DPMs have been applied to various applications, including image generation~\citep{sohl2015deep,song2020score,ho2020denoising}, image editing~\citep{avrahami2022blended,choi2021ilvr}, 3D model generation~\citep{Luo_2021_CVPR,lyu2023controllable,liu2023meshdiffusion,zeng2022lion,nichol2022pointe}, and conditional generation~\citep{choi2021ilvr}, demonstrating significant performance in the generative domain.
%
Considering the {\it diverse} categories of dental malocclusion in clinical settings and the difficulty of data collection, we model the automatic tooth arrangement task as {\it a generative task}. We introduce DPMs to generate tooth transformation matrices, which are expected to learn the orthodontic data distribution, thereby increasing the generalization of the network and enabling more effective handling of real clinical data.


\begin{figure*}[t]
    \centering
    \includegraphics[width=1\linewidth]{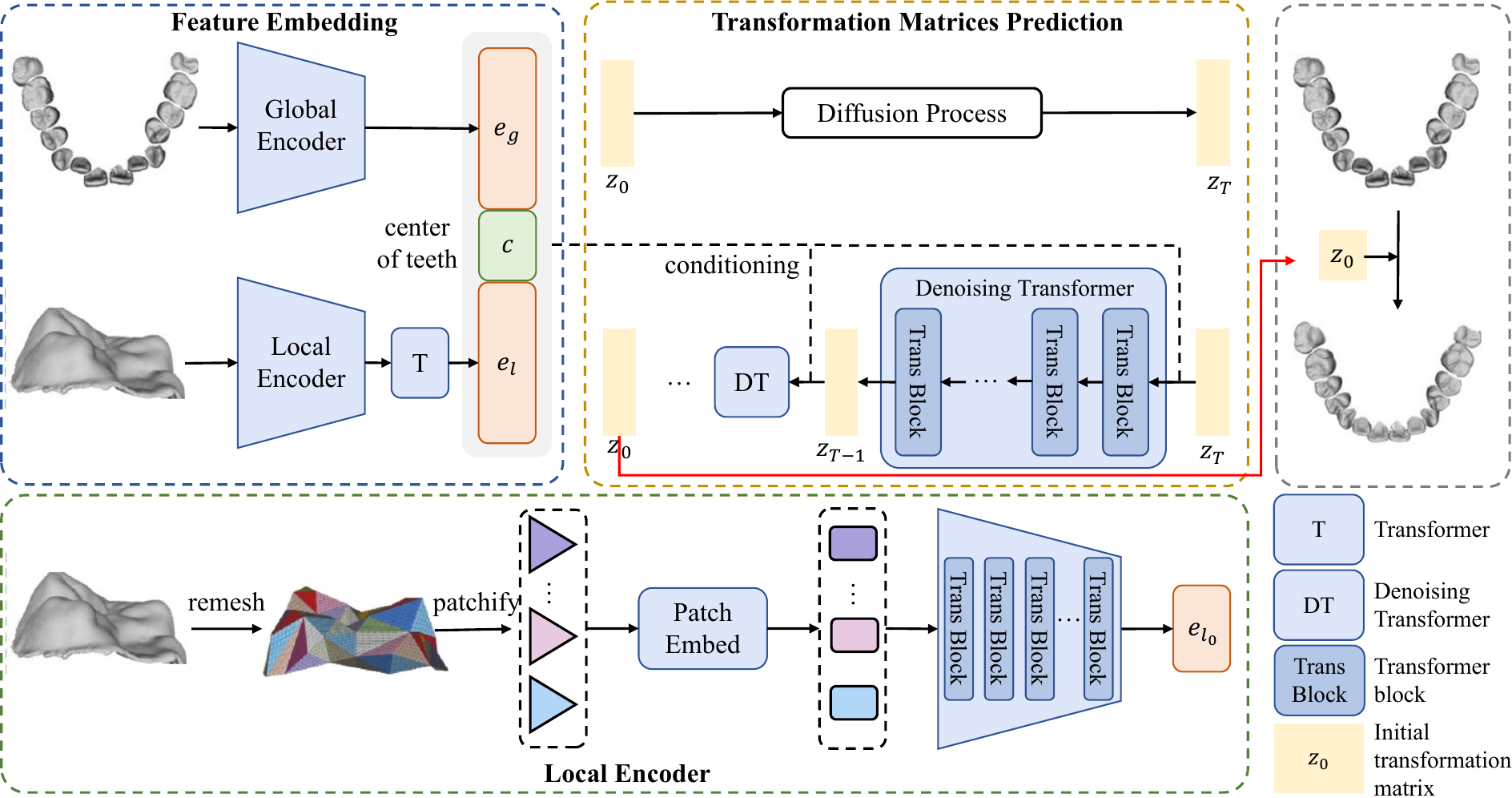}
    \caption{The overall architecture of the proposed framework. In the feature embedding module, we utilize MeshMAE \citep{liang2022meshmae} as the local encoder network, where the faces of the single tooth mesh are divided into patches and embedded to compute the feature embedding $e_{l_0}$, then $e_{l_0}$ is fed into a transformer network to produce the local feature $e_{l}$. We utilize PointNet++ as the global encoder network, where the point cloud of the whole jaw is utilized to extract the global feature $e_g$.
    In the transformation matrices prediction module, we build upon diffusion models and design 
    the denoising transformer to predict the transformation matrices conditioned on the extracted local and global features.
    The lower part of the figure presents the local encoder, including the data pre-processing and feature embedding process. }
    \label{fig:framework}
\end{figure*}
\section{Methods}
\label{section:3}

\subsection{Overview}

The tooth arrangement task aims at estimating a 6-dimensional transformation parameter for each tooth in the input dental models.
As shown in~\cref{fig:framework}, we propose TADPM to solve this task, comprising two key modules:
a Feature Embedding module and a Transformation Matrices Prediction module. 
Specifically, we denote the input data, 
a set of segmented dental mesh models (without gingiva),
as $M=\{M_k \vert k \in \mathcal{K}\}$,
where $\mathcal{K}$ is the set of teeth labels according to the FDI two-digit notation~\citep{harris2005tooth}, and $M_k= (V_k, F_k)$ denotes the 3D mesh model of a single tooth with label $k$, where $V_k$ and $F_k$ are the collections of vertices and faces of $M_k$, respectively. 
First, the Feature Embedding module uses a local encoder and a global encoder to separately extract the local and global features from the input dental models. 
Then, 
the Transformation Matrices Prediction module  
generates the pose transformation matrix for each tooth conditioned on the feature embeddings with the Denoising Transformer blocks. We describe these two modules in detail as follows.

\subsection{Feature Embedding Module}
\label{ssec: feature_embedding}



Existing tooth arrangement methods~\citep{wei2020tanet,li2020malocclusion,lingchen2020iorthopredictor,wang2022tooth} 
only use point clouds as input to extract features. 
Compared with point clouds, 3D mesh models contain more geometric details (especially the topological connectivity) and can provide fine geometric features that are helpful for subsequent transformation prediction; however, mesh processing requires more time and resources. 
To achieve 
effective and efficient feature extraction of dental models, we propose to use both 3D meshes and point clouds for feature extraction. This method involves extracting local features from 3D meshes to accurately represent the shape of individual teeth, and global features from point clouds to conserve computational time and memory.




\subsubsection{Local Feature}



%
 
In this paper, we conduct tooth arrangement on dental models in the 3D mesh representation.
Compared with point clouds,
mesh models contain rich connectivity-related information, 
which inspires us to extract powerful features directly from the mesh models.
However, due to the intrinsic irregular nature of the data format of mesh models and complex adjacency between vertices, it is difficult to directly use neural networks to process 3D meshes. 
Moreover, due to the difficulty of annotating labels for dental models, the number of training samples is actually insufficient for performing supervised learning on dental meshes.
To address these limitations, we introduce MeshMAE~\citep{liang2022meshmae} to extract the feature embedding of each tooth, so that the self-supervised pre-training process in MeshMAE would benefit the downstream tooth arrangement tasks.   
 
First, following MeshMAE, we perform the re-mesh operation on a single tooth mesh to construct a hierarchical structure. During the process, the faces $F_k$ in mesh $M_k$ are divided into several non-overlapping patches.
Then, we concatenate the features of all faces in a patch into the patch feature according to the order from the re-meshing process. 
The patch feature is then projected into the patch embedding by a multi-layer perceptron (MLP).
Furthermore, we integrate positional information by computing the positional embedding based on the center coordinates of the patches, which is more appropriate for unordered geometric data.
Finally, the patch embedding and positional embedding are combined to form the input embedding.

To address the challenge posed by  
limited data, we adopt the self-supervised pre-training strategy of MeshMAE to learn an effective feature representation from the re-meshed dental datasets. Specifically, a portion of the input embeddings are randomly masked, and only the remaining unmasked embeddings are fed into the encoder for further processing. Then, the masked embeddings are replaced with a shared mask embedding, which are eventually combined with the encoder's output to produce the input of the decoder. Finally, the decoder predicts the face 
features and vertex 
coordinates of the masked patches to reconstruct
the information of the masked parts. 
After the pre-training stage, we utilize the encoder of MeshMAE as the local encoder 
and finetune\footnote{See the training process in \cref{alg:train}.} it to compute the feature embedding  $e_{l_{k_0}}$ of the tooth $M_k$.




\subsubsection{Feature Propagation}
The local features extracted by MeshMAE capture the tooth-level geometric details. However, they lack the interdependent information shared among adjacent teeth.
To address this issue, we introduce a Transformer network to transfer geometric information among all teeth. The output sequence $\{e_{l_{k_0}} \vert k \in \mathcal{K}\}$ obtained from the local encoder is fed into the Transformer, and 
the geometry center $c_k$ of $M_k$  is utilized as the positional embedding. By leveraging the attention mechanism, the resulting output $\{e_{l_{k}} \vert k \in \mathcal{K}\}$ is able to capture
inter-teeth mutual information across all teeth.
This mechanism enables the model to effectively capture and integrate the context of geometry relationships among all teeth.

\subsubsection{Global Feature}


Instead of using mesh as in the local feature extraction, we propose to utilize point clouds to extract the global feature of the whole dental model. The reasons are two-fold:
1) on the one hand, using MeshMAE to extract global features is memory-consuming since there are too many faces in the whole jaw;
2) on the other hand, local features already contain sufficient fine geometric details,  
which makes it redundant to re-extract geometric details for global feature extraction.
Therefore, we leverage point clouds for the extraction of global features from the entire dental model. 

Specifically, we define the input point cloud as $P=\{p_k \in \mathbb{R}^{N_k \times 3}|k\in \mathcal{K}\}$, where $p_k$ denotes the point cloud sampled from the single tooth mesh $M_k$ and $N_k$ denotes the number of points in $p_k$. Then, we use PointNet++~\citep{qi2017pointnet} as the global feature extraction module to obtain a global feature $e_g$ from $P$. Although features extracted by PointNet++ do not contain very fine geometry details, they nevertheless provide sufficient global information for tooth arrangement. 

Finally, the local and global features and the center of teeth are integrated through Stack and Concatenation operations to form the overall feature representation of the dental model:
\begin{equation}
      e = \mathrm{Stack}\{\mathrm{Concat}(e_{g},c_k, e_{l_k})\vert k\in \mathcal{K}\},
\end{equation}
where $\mathrm{Stack}$ and $\mathrm{Concat}$ denote corresponding operations in PyTorch respectively.  

\subsection{Transformation Matrices Prediction Module}


After feature extraction, the tooth arrangement task can be regarded as a generation task conditioned on the input feature $e$. We adopt a representative diffusion model~\citep{ho2020denoising} by elaborately reformulating our task as a {\it conditioned} DPM model to generate a 6-DoF transformation matrix of each tooth.

During the training stage, each tooth is assigned with a 6-DoF ground truth\footnote{Note that we construct the training dataset by shifting dental models after orthodontics, so we can obtain the ground truth transformation matrices directly.} transformation parameter $z_0^k = (m_0^k, r_0^k)$, where $m_0^k \in \mathbb{R}^3$ and $r_0^k \in \mathbb{R}^3$ respectively denote the translation and rotation vectors of the single tooth $M_k$ in the coordinate system. By concatenating all $\{z_0^k | k\in \mathcal{K}\}$
, we construct the transformation matrix $z_0 \in \mathbb{R}^{|\mathcal{K}|\times 6}$, which serves as input for the diffusion model, as illustrated in~\cref{fig:framework}. Then, we introduce perturbations to $z_0$ to generate a noisy matrix $z_t$ at timestep $t$:
\begin{equation}
      z_t = \alpha_tz_0 + \sigma_t\boldsymbol\epsilon,
\end{equation}
where $\boldsymbol\epsilon \sim \mathcal{N}(\mathbf{0}, \mathbf{I})$,  $\alpha_t$ and $\sigma_t$ are as defined in~\cref{eq:forward}.

Subsequently, we employ a neural network, denoted as $z_\theta(z_t, t, e)$, to predict the original $z_0$ from the noisy $z_t$ with the feature $e$ extracted from the feature embedding module as the condition.
The network's actual output, $\bar{z}_0 = \{\bar{z}_0^k\vert k \in \mathcal{K}\} \in \mathbb{R}^{|\mathcal{K}|\times 6}$, is a sequence of 6-DoF transformation parameters.
Each $\bar{z}_0^k = (m_0^k, r_0^k) \in \mathbb{R}^6$ defines the transformation parameters of the single tooth mesh $M_k$, where $m_0^k$ and $r_0^k$ represent the translation and rotation parameters of $M_k$, respectively.  We can obtain a transformation matrix $T_k\in \mathbb{R}^{4\times4}$ from $z_0^k$ through SE(3) exponential mapping, defined as follows:
\begin{equation}
      \Omega = \left [ \begin{matrix}
        0& -r_k^z& r_k^y \\
        r_k^z& 0& -r_k^x \\
        -r_k^y& r_k^x& 0 \\
        \end{matrix} \right ]
\end{equation}

\begin{equation}
      R_k = exp(\Omega) = I_{3\times3} + \frac{sin\theta}{\theta}\Omega + \frac{1-cos\theta}{\theta^2}\Omega^2
\end{equation}
\begin{equation}
     T_k = \left [ \begin{matrix}
        R_k & m_0^k \\
        \textbf{0} & 1\\
        \end{matrix} \right ]
\end{equation}

Finally, we apply the transformation matrix $T_k$ on the original dental model $M$ to get an aligned dental model $\bar{M}$. For convenience, we represent the whole aligning process as follows:
\begin{equation}
      \bar{M} = \emph{aligner}(M,\bar{z}_0).
\end{equation}


\renewcommand{\algorithmicrequire}{\textbf{Input:}}
\renewcommand{\algorithmicensure}{\textbf{Output:}}
\begin{algorithm}[!ht]
  \caption{Training stage of TADPM }\label{alg:train}
  \label{alg1}
  \small
  \begin{algorithmic}[1]
    \REQUIRE Features $e$, Ground truth of transformation matrix $z_0$, dental model $M$ before orthodontic treatment and ground truth $M^*$ after treatment.
    \REPEAT
    \STATE $t \sim \mathrm{Uniform}[1,\cdots,T]$
    \STATE $\boldsymbol\epsilon \sim \mathcal{N}(\mathbf 0,\mathbf I)$
    \STATE $z_t \leftarrow \alpha_tz_0 + \sigma_t\boldsymbol\epsilon$
    \STATE $\bar{z}_0 \leftarrow z_\theta(z_t,t,e)$
    \STATE $\bar{M} \leftarrow \emph{aligner}(M,\bar{z}_0)$  
    \STATE Take gradient descent step on \\
    $$\nabla_\theta [\lambda_1 \mathcal{L}_{CD} + \lambda_2 \mathcal{L}_{diff} + \lambda_3 \mathcal{L}_{pos}] $$
	\UNTIL converge
  \end{algorithmic}
\end{algorithm}

The training process is summarized in \cref{alg1}, where step 7 refers to the loss function that we describe in Sec.~\ref{ssec: loss}.
In the sampling stage, we first generate a random transformation matrix $z_T \sim \mathcal{N}(\mathbf 0,\mathbf I)$ and gradually denoise it to predict the desired transformation parameters of each tooth. We further adopt the determinative sampling method introduced in ~\citep{song2020denoising} to accelerate the sampling process:
\begin{equation}
      z_{t-\Delta t} = \sqrt{\bar{\alpha}_{t-\Delta t}}\bar{z}_0 + \sqrt{1-\bar{\alpha}_{t-\Delta t}}\frac{z_{t}-\sqrt{\bar{\alpha}_t}\bar{z}_0}{\sqrt{1-\bar{\alpha}_t}},
\end{equation}
where $\Delta t$ denotes the step size. \cref{alg2} summarizes the whole sampling process.

\begin{algorithm}[t]
    \caption{Sampling stage of TADPM}\label{alg:inference}
    \label{alg2}
    \small
    \begin{algorithmic}[1]
      \REQUIRE Features $e$.
      \ENSURE Prediction of transformation matrix $z_0$.
      \STATE $z_T \sim \mathcal{N}(\mathbf 0,\mathbf I)$
      \FOR{$t=T,T-\Delta t,\cdots,\Delta t$}
        \STATE $\bar{z}_0 \leftarrow z_\theta(z_t,t,e)$
        \STATE $z_{t-\Delta t} \leftarrow \sqrt{\bar{\alpha}_{t-\Delta t}}\bar{z}_0 + \sqrt{1-\bar{\alpha}_{t-\Delta t}}\frac{z_{t}-\sqrt{\bar{\alpha}_t}\bar{z}_0}{\sqrt{1-\bar{\alpha}_t}}$
      \ENDFOR
      \RETURN $z_0$
    \end{algorithmic}
  \end{algorithm}

\subsection{Loss Functions}
\label{ssec: loss}
\subsubsection{Reconstruction Loss}
Chamfer distance is a 
widely used
metric for point cloud reconstruction tasks. 
We first convert the predicted result $\bar{M}$ and ground truth $M^*$ to point clouds by simply utilizing their vertex sets $\bar{V}$ and $V^*$. 
Then, we compute the chamfer distance between corresponding teeth in the prediction result $\bar{M}$ and the ground truth $M^*$:
%
\begin{equation}
\mathcal{L}_{CD} = \sum_{k\in \mathcal{K}}(\frac{1}{\vert\bar{V}_k\vert}\sum_{x \in \bar{V}_k} \min_{y \in V^*_k} \Vert x-y\Vert^2 + \frac{1}{\vert V^*_k\vert}\sum_{y \in V_k^*} \min_{x \in \bar{V}_k} \Vert x-y \Vert^2),
\end{equation}
where $\bar{V}_k$  denotes vertices of the predicted tooth with label $k$ and $V^*_k$ denotes vertices of the ground truth with label $k$. $\mathcal{K}$ is the set of labels for all teeth.

\subsubsection{Diffusion Loss}
We use the predicted transformation matrix output by diffusion model to supervise the training directly. Generally speaking, the squared error between prediction and uncorrupted data is a natural choice for the loss function of diffusion models:
\begin{equation}
      \mathcal{L}_{diff} = \Vert z_0 -z_\theta(z_t,t,e) \Vert^2,
\end{equation}
where $z_0$ and $z_\theta(z_t,t,e)$ denote the ground truth and the predicted transformation matrix, respectively. 

\subsubsection{Relative Position Loss}
Relative positional relationship among teeth constrains the space between adjacent teeth and is critical to tooth arrangement. 
We define a distance matrix $D$ to represent the positional structure of a dental model $M$:
\begin{equation}
     D(M)_{ij} := \Vert c_i - c_j \Vert_1,
\end{equation}
where $D(M)_{ij}$ denotes the distance between the $i$-th and $j$-th teeth, and $c_i$ and $c_j$ denote the geometry center of the $i$-th and $j$-th tooth, respectively.
Compared with the aforementioned chamfer distance, calculating our designed distance matrix requires less time.

We then calculate the difference between the distance matrices of the prediction result and the ground truth, providing supervision for the relative positional relationship:
\begin{equation}
     \mathcal{L}_{pos} = \Vert D(\bar{M}) - D(M^*) \Vert^2,
\end{equation}
where $D(\bar{M})$ and $D(M^*)$ denote the distance matrices of the prediction result $\bar{M}$ and the ground truth $M^*$, respectively.

$\mathcal{L}_{pos}$ constrains distances between corresponding teeth in the upper and lower jaws as well as distances between adjacent teeth, thereby achieving a better occlusal relationship and tooth arrangement.

Finally, we combine the above three losses together to obtain the final loss function:
\begin{equation}
     \mathcal{L} = \lambda_1 \mathcal{L}_{CD} + \lambda_2 \mathcal{L}_{diff} + \lambda_3 \mathcal{L}_{pos},
\end{equation}
where $\lambda_1$, $\lambda_2$, and $\lambda_3$ denote weights for each part of the loss function, and they are set to 0.05, 0.5, and 1 in our experiments.

\section{Experimental Results}

Since there is no public dataset for tooth arrangement available, we evaluated the proposed TADPM on our newly built dataset collected from clinical patients. We compared TADPM with several baselines qualitatively and quantitatively.
We also conducted ablation studies to demonstrate the effectiveness of each component in TADPM.

\subsection{Experimental Setups}

\subsubsection{Datasets}%
Our dataset\footnote{
The amount of dental models in our dataset is comparable to that of the datasets in existing methods, e.g., the dataset in  ~\citep{li2020malocclusion} has 178 pairs of dental models.
} consists of 212 pairs of dental models before and after orthodontic treatment, which are collected from patients treated between June 2016 and April 2023. 
We invited four orthodontists with over ten years of clinical experience to annotate the tooth segmentation labels and tooth position numbering. The doctors were trained to use the `Mesh Labeler' software and delineated the boundaries of the teeth according to the FDI international tooth numbering system. 

In our experiment, we observed that some of the tooth segmentation labels are different before and after orthodontic treatment in our dataset (see~\cref{fig:single} for an example), which would have a negative impact on the training process. To address this problem, we added random disturbance to the dental models after orthodontics to simulate ``pre-orthodontic'' dental models.  For each patient, we randomly created 10 pairs. The constructed data pairs are used for {\it training}, while the real data pairs are used for {\it testing}.
Additionally, the dental models in our dataset are usually distributed far from the origin, leading to challenges in feature extraction. To mitigate this issue, we relocated the geometric centers of all dental models to the origin. 

\subsubsection{Evaluation Metrics}
Following~\citep{wei2020tanet,li2020malocclusion, wang2022tooth}, we adopt the metrics of ADD~\citep{hinterstoisser2013model}, PA-ADD, cosine similarity accuracy (CSA), and $\mathrm{ME}_{rot}$ to measure the performance of our method. ADD is the point-wise mean distance between the predicted dental models and ground truth, which provides an intuitive measure of alignment error. 
PA-ADD is the ADD metric calculated after rigid registration from the prediction to the ground truth.
Since the process of tooth arrangement can be described by predicting the transformation parameters of each single tooth, 
we can calculate the CSA of predicted matrices and the ground truth. $\mathrm{ME}_{rot}$ denotes the mean error of rotation.



\begin{figure}[t]
    \centering
\includegraphics[width=0.7\linewidth]{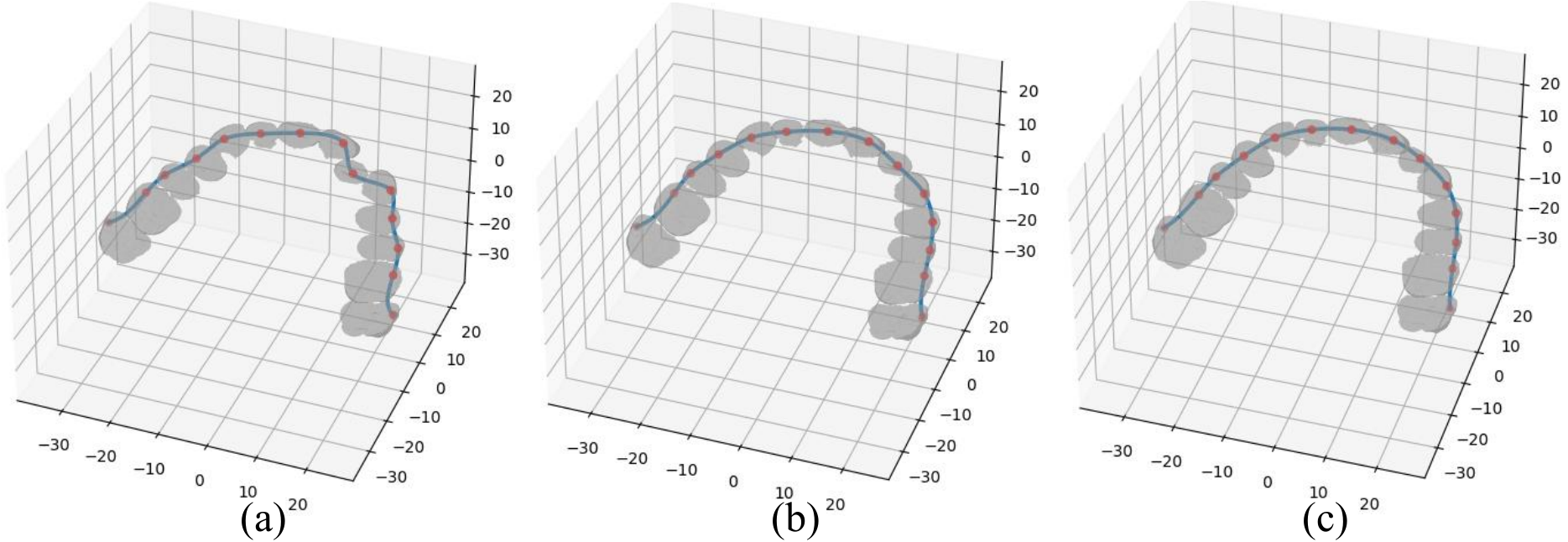}
\caption{Visualization of the dental arch curves, where (a), (b), and (c) show the arch curve of input data, prediction, and ground truth, respectively. The coordinate unit is mm.}
\label{fig:curve}
\end{figure}

To measure the teeth alignment effect and occlusion relationship between dental arches, we introduce the Fréchet Distance~\citep{eiter1994computing} between dental arch curves of prediction and ground truth as a metric, denoted as $\mathrm{FD}_{cur}$. We leverage the B-spline curve to represent the dental arch curve, which is obtained by interpolating the landmarks (red points in the figure) on the teeth, as shown in  \cref{fig:curve}.

\subsubsection{Implementation Details}
We train the proposed TADPM on the platform of PyTorch~\citep{paszke2019pytorch}, in a Linux environment with 2 NVIDIA Tesla A100 GPUs. In all our experiments, we set the batch size to 8 and the epochs to 500. We use the AdamW optimizer with a learning rate of 1e-4. We use 12 transformer blocks for the pertaining and also for the diffusion model. 

The number of parameters in our model is 173M, and the computational complexity is approximately 6,000 GFlops. During the inference stage, we adopt the DDIM sampler for acceleration. The average inference time of a single dental model is 38.72s, which is acceptable for clinical usage.

\subsection{Comparisons with State-of-the-Arts}

To demonstrate the effectiveness of TADPM, we compare it with three state-of-the-art methods, i.e. TANet~\citep{wei2020tanet}, PSTN~\citep{li2020malocclusion}, and TAligNet~\citep{lingchen2020iorthopredictor}. The experimental results are summarized in Table~\ref{table:1}, where our method achieves the best performance on all four metrics. Notably, our method shows significant advantages on ADD and $\mathrm{FD}_{cur}$, demonstrating that our method can not only obtain good tooth alignment effectiveness but also improve the occlusal relationship of the dental arches.


\begin{table}[!t]
\caption{Comparison with state-of-the-art methods. $\downarrow$ indicates the lower the better, while $\uparrow$ indicates the higher the better. The coordinate unit is degree for $\mathrm{ME}_{rot}$, mm for ADD, PA-ADD and $\mathrm{FD}_{cur}$.}
\label{table:1}
\centering 
\fontsize{9pt}{11pt}\selectfont
\begin{tabular}{ccccccc}
\hline
 Model  & ADD ($\downarrow$)  &  PA-ADD ($\downarrow$)  &  CSA  ($\uparrow$)  &  $\mathrm{ME}_{rot}$  ($\downarrow$) & $\mathrm{FD}_{cur}$  ($\downarrow$) \\ 
\hline
TANet & 1.618 & 1.325 & 0.813 & 5.564 & 2.320\\ \hline 
PSTN & 1.784 & 1.440 & 0.781  & 6.073 & 2.552\\\hline 
TAligNet & 1.752 & 1.418 & 0.785 & 5.893 & 2.665\\\hline
TADPM (Ours) & \textbf{1.487} & \textbf{1.202} & \textbf{0.847} & \textbf{5.472} & \textbf{1.932}\\
\hline
\end{tabular}
\end{table}

\begin{figure}[!t]
    \centering
\includegraphics[width=0.5\linewidth]{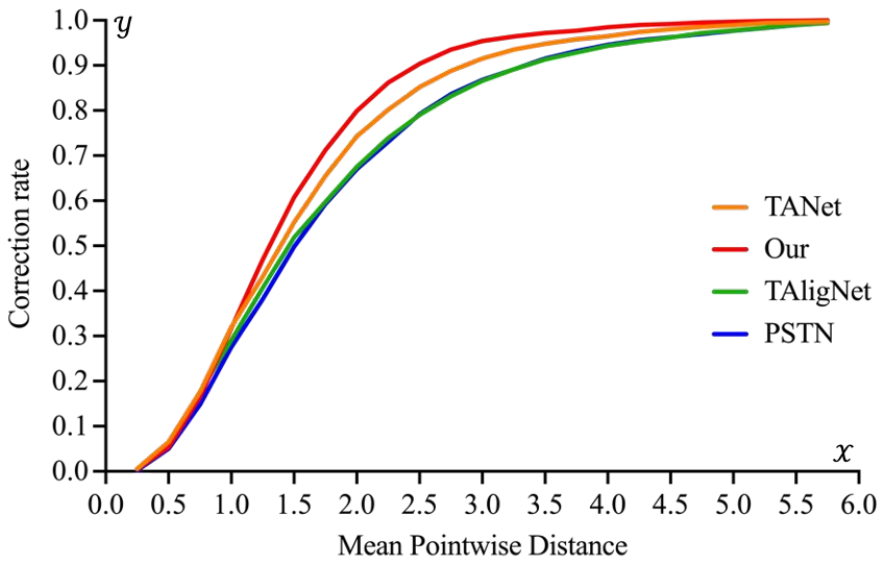}
    \caption{
    The distribution of pointwise distance, where the 
    $x$-axis
    represents the range of the mean pointwise distance, and the 
    $y$-axis
    represents the accuracy within the range.}
    \label{fig:linechart}
\end{figure}

In addition, we show the distribution of mean pointwise distance (the average of distances between each pair of corresponding vertices in predicted tooth mesh and ground truth)  in \cref{fig:linechart}, where the $x$-axis represents the range of mean pointwise distance and the $y$-axis
represents the accuracy within this range. Compared with baseline methods, the results of our method have a wider distribution within a smaller range, indicating that the tooth arrangement effect of our method is closer to the ground truth.

\subsection{Qualitative Evaluation}

\begin{figure*}[t]
    \centering
\includegraphics[width=1\linewidth]{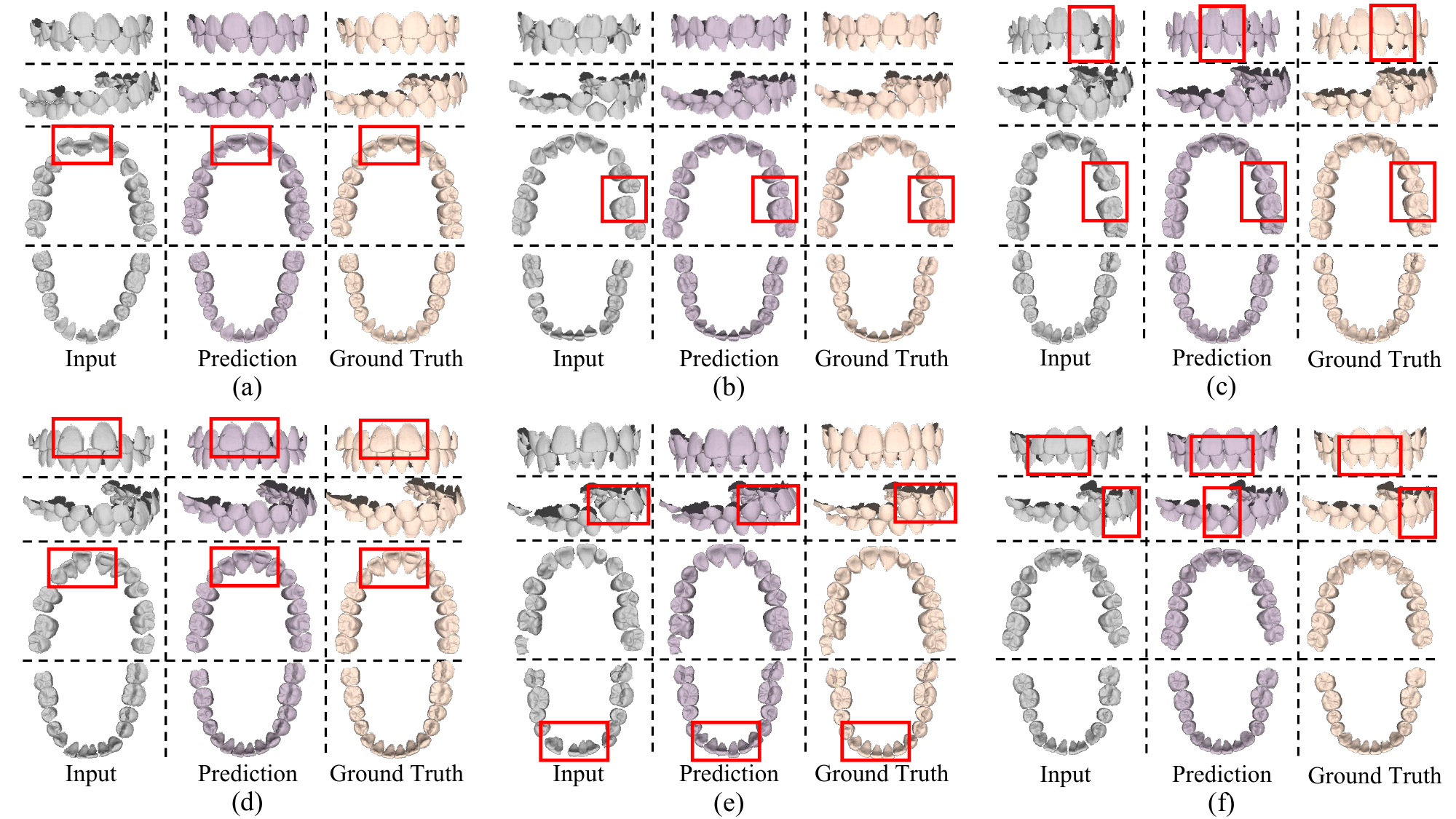}
\caption{Visualization of the orthodontic results of our method, where (a) - (d) belong to the malocclusion problem of teeth crowding, (e) belongs to the malocclusion problems of diastema and deep overjet, and (f) belongs to the malocclusion problem of deep overbite. We demonstrate more predicted orthodontic results in supplemental materials.
}
\label{fig:sample}
\end{figure*}

\cref{fig:sample} illustrates some examples of the aligned dental models obtained by our method, which can generate satisfactory dental alignment effects for different kinds of malocclusions, demonstrating the generalization of our method.
In detail, our method focuses not only on tooth alignment issues such as teeth crowding or diastema but also on dental occlusion problems such as deep overbite or deep overjet. Tailored to distinct problems, our method ensures even and smooth teeth arrangements. 


 \subsection{Subjective Evaluations}


In this subsection, we provide subjective evaluations of dental alignment effectiveness from professional clinical orthodontists. The responses were collected from 5 orthodontists, who were asked to rate 10 randomly selected samples by answering three questions for each sample:
\begin{itemize}
    \item Question I: Does the alignment result of the upper and lower jaw meet the clinical standards? 
    \item Question II: Does the occlusal relationship between the upper and lower jaw meet the clinical standards? 
    \item Question III: Does the result meet your expectation for the post-treatment effects of the patient? 
\end{itemize}

For each question, orthodontists are required to rate between 1 and 5 scores, with 5 representing the best. We list the average and variance of rates for each sample in Table~\ref{table:4}. The average rates for three questions are all over 4 (Good), indicating that the results of TADPM can meet clinical standards.


\begin{table*}[!ht]
\caption{Orthodontists' rates for 10 randomly selected samples.}
\label{table:4}
\centering
\fontsize{9pt}{11pt}\selectfont
\begin{tabular}{cc| c|c|c|c|c|c}
\hline
\multicolumn{2}{c}{\multirow{2}{*}{Model}}&  \multicolumn{2}{|c}{Question I} & \multicolumn{2}{|c}{Question II} & \multicolumn{2}{|c}{Question III}   \\
\cline{3-8}
\multicolumn{2}{c|}{} & Average & Variance & Average & Variance & Average & Variance \\
\hline
\multicolumn{2}{c|}{Sample I}&4.20&0.16&4.80&0.16&4.40&0.24\\
\multicolumn{2}{c|}{Sample II}&3.80&0.56&4.20&0.16&4.00&0.40\\
\multicolumn{2}{c|}{Sample III}&4.20&0.16&4.60&0.24&4.20&0.16\\
\multicolumn{2}{c|}{Sample IV}&3.80&0.56&4.80&0.16&4.40&0.64\\
\multicolumn{2}{c|}{Sample V}&4.40&0.24&4.60&0.24&4.60&0.24\\
\multicolumn{2}{c|}{Sample VI}&4.00&0.00&4.40&0.24&4.20&0.16\\
\multicolumn{2}{c|}{Sample VII}&4.00&0.00&4.40&0.24&4.40&0.24\\
\multicolumn{2}{c|}{Sample VIII}&4.40&0.24&4.80&0.16&4.20&0.16\\
\multicolumn{2}{c|}{Sample IX}&4.20&0.16&4.60&0.24&4.20&0.16\\
\multicolumn{2}{c|}{Sample X}&3.60&0.24&4.40&0.64&4.20&0.16\\
\hline
\multicolumn{2}{c|}{Average}&4.06&0.23&4.56&0.25&4.28&0.26 \\
\hline
\end{tabular}
\end{table*}






\subsection{Ablation Study}

To further analyze the effectiveness of the components and loss functions of the proposed method, we conduct ablation studies on the proposed modules and loss functions.





\subsubsection{Effectiveness of Proposed Modules}


As described in Sec.~\ref{section:3}, we adopt the MeshMAE as the local encoder and introduce the diffusion probabilistic module to predict the transformation matrix. To verify the effectiveness of the above two strategies, we conduct ablation studies by replacing the MeshMAE and the diffusion probabilistic module with PointNet++ and MLP, respectively.  
The experimental results are shown in columns 2 (``w/o MeshMAE'') and 5 (``w/o DPM'') in Table~\ref{table:2}.
To promote the feature extraction effect, we introduce the global encoder and the feature propagation module into our model. Here, we conduct the ablation experiments by removing these two parts to verify whether they are necessary for feature extraction. The experimental results are shown in columns 3  (``w/o Global'') and 4 (``w/o FP'') in Table~\ref{table:2}.

The results of our full method are shown in the last column of Table~\ref{table:2}. From these results, the appropriate feature embedding modules can significantly improve the performance. Unlike other methods that utilize MLP to directly regress the transformation matrices, we introduce the diffusion model to learn the distribution of transformation matrices. This strategy further enhances the effectiveness of our method on all metrics, especially on $\mathrm{FD}_{cur}$, demonstrating that the introduction of the diffusion model can indeed bring a good occlusal relationship.

\begin{table}[!t]
\caption{The effectiveness of different modules. ``Local'' denotes the local encoder module, ``Global'' denotes the global encoder module, ``FP'' denotes the feature propagation module, ``DPM'' denotes the diffusion probabilistic module, and the last column is the result of our method.}
\label{table:2}
\centering
\fontsize{9pt}{11pt}\selectfont
\begin{tabular}{c|c|c|c|c|c}
\hline
Modules& w/o MeshMAE& w/o Global &w/o FP & w/o DPM & Ours \\\hline
Local &PointNet++&\checkmark&\checkmark&\checkmark&\checkmark\\
Global &\checkmark&&\checkmark&\checkmark&\checkmark\\
FP&\checkmark&\checkmark&&\checkmark&\checkmark\\
DPM&\checkmark&\checkmark&\checkmark&MLP&\checkmark\\\hline

 ADD ($\downarrow$)&1.637&1.793&1.703&1.536&\textbf{1.487}\\
 PA-ADD ($\downarrow$)&1.336&1.507&1.430&1.248&\textbf{1.202}\\
 CSA ($\uparrow$)&0.813&0.770&0.792&0.836&\textbf{0.847}\\
 $\mathrm{ME}_{rot}$ ($\downarrow$)&5.652&5.951&5.695&5.492&\textbf{5.472}\\
 $\mathrm{FD}_{cur}$ ($\downarrow$)&2.638&3.073&2.607& 2.152 & \textbf{1.932} \\
\hline
\end{tabular}
\end{table}

\subsubsection{Ablation Study on Loss Functions}

The loss functions in our full method include three parts, i.e., reconstruction loss $\mathcal{L}_{CD}$, diffusion loss $\mathcal{L}_{diff}$, and relative position loss $\mathcal{L}_{pos}$. We conduct ablation studies to verify the effectiveness of each part. The results are shown in Table~\ref{table:3}. We can observe that it is difficult to achieve good results using only $\mathcal{L}_{diff}$ to guide model learning due to the lack of geometric information, while the combination of $\mathcal{L}_{diff}$ and $\mathcal{L}_{CD}$ can boost the performance. To constrain the distance between adjacent teeth, we also add $\mathcal{L}_{pos}$ to the overall loss function, which contributes to achieving better results.

\begin{table}[!t]
\caption{The performance of different loss functions, where the last row is the result of our method.}
\label{table:3}
\centering
\fontsize{9pt}{11pt}\selectfont
\begin{tabular}{c|ccccc}
\hline
$\mathcal{L}$& ADD ($\downarrow$)  &  PA-ADD ($\downarrow$)  &  CSA  ($\uparrow$)  &  $\mathrm{ME}_{rot}$  ($\downarrow$)&$\mathrm{FD}_{cur}$  ($\downarrow$)\\
\hline
$\mathcal{L}_{CD}$&1.587&1.319&0.829&5.504&2.034\\
$\mathcal{L}_{diff}$&1.709&1.531&0.797&5.879&2.891\\
$\mathcal{L}_{CD}+\mathcal{L}_{diff}$&1.502&1.229&0.832&5.478&\textbf{1.879}\\
$\mathcal{L}_{CD}+\mathcal{L}_{diff}+\mathcal{L}_{pos}$&\textbf{1.487}&\textbf{1.202}&\textbf{0.847}&\textbf{5.472}&1.932\\ \hline
\end{tabular}
\end{table}

\subsection{Iterative Experiment}

\begin{figure*}[ht]
    \centering
\includegraphics[width=0.9\linewidth]{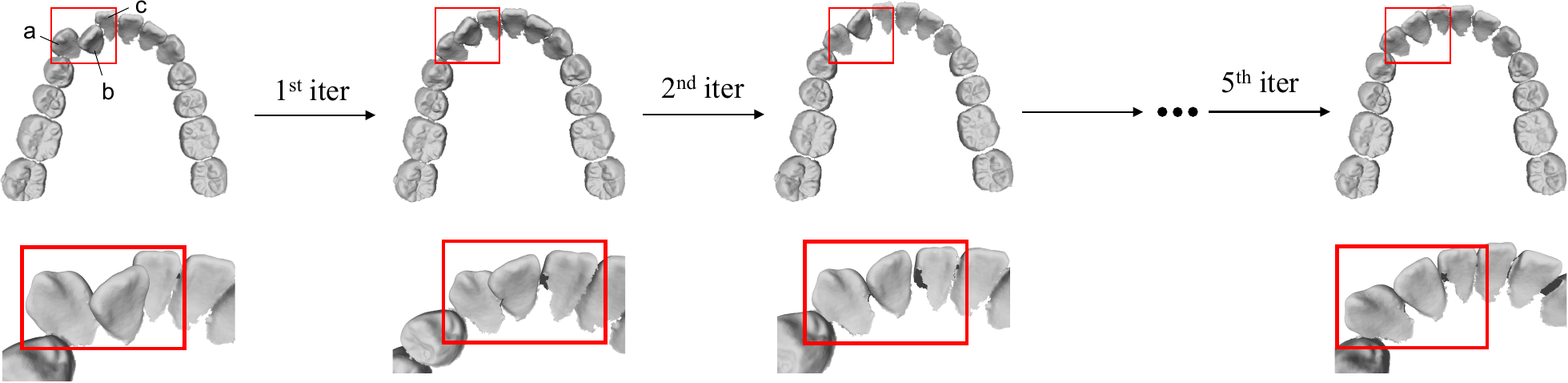}
    \caption{ 
    An example of the iterative experiment result. The second row shows the details of the red boxes in the first row. }
    \label{fig:iter}
\end{figure*}

The output of our method is a set of teeth, which is in the same format as the input data. Therefore, a natural assumption is that the network can achieve better tooth alignment than the current output by reusing the output as input.
Based on this assumption, we explore the performance of iterative experiments, i.e., using the output of our network as input and conducting teeth alignment iteratively.
~\cref{fig:iter} shows an example of the iterative experiment. We can observe from the original input that the space between teeth $a$ and $c$ is too small to let tooth $b$ fit in. After the first iteration, $a$ and $b$ are touching. In the subsequent iterations, the network gradually adjusts the whole dental model towards a global optimization.

Iterative teeth alignment is helpful in extreme cases like~\cref{fig:iter}, where the network fails to align all teeth well at once.
But in most cases, our model can already achieve good performance in one round.


\section{Conclusion}

In this paper, we present a novel automatic tooth arrangement network to generate tooth transformation matrices for orthodontics treatment. To address the challenges posed by complex real-patient dental data, our method constructs the ``pre-orthodontic'' teeth dataset by randomly moving the teeth after orthodontic treatment and leverages the diffusion probabilistic models to generate the transformation parameters for each tooth. Additionally, we extract local geometric details of the tooth based on mesh models, which is verified to contribute to the effectiveness of feature embedding extraction in our experiments. Extensive experiments and user studies demonstrate the effectiveness of our method.





\bibliography{mybibfile}

\end{document}